Politechnika Warszawska
Wydział Elektroniki i Technik Informacyjnych
Instytut Informatyki

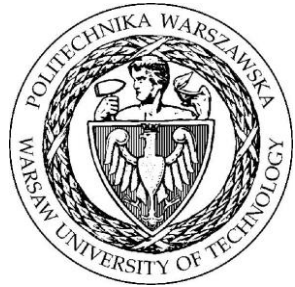

Studia Podyplomowe Analityka Biznesowa
PRACA KOŃCOWA

mgr Karol Chlasta

# Model analizy wydźwięku na Twitterze dla języka polskiego

Opiekun pracy
prof. nadzw. dr hab. inż. Piotr Gawrysiak

Warszawa, 2015

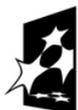
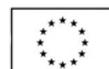

# STRESZCZENIE

Streszczenie pracy w języku polskim.

Analiza text-minigowa danych twitterowych w związku z wyborami Prezydenta Polski ogłoszonymi na dzień 10 maja 2015 roku. W ramach pracy zaimplementowano silnik pobierania danych z twittera, zbudowano i oczyszczono korpus tekstowy oraz stworzono macierz TDM. Każdy tweet z korpusu tekstowego oceniono ze względu na jego wydźwięk na podstawie ilości występujących nim emotikonów oraz polskich słów o pozytywnym lub negatywnym wydźwięku. Tak przygotowany zbiór danych został wykorzystany do zbudowania, trenowania i testowania czterech różnych algorytmów uczenia maszynowego, w celu wybrania najbardziej odpowiedniego do automatycznej klasyfikacji nowych tweetów w przyszłości. Najlepszą dokładność klasyfikacji uzyskano przy pomocy naiwnego klasyfikatora bayesowskiego i przy wykorzystaniu metody maksymalnej entropii, odpowiednio 71.76% oraz 77.32%. Całość rozwiązania wykonano przy użyciu języka programowania R.

Słowa kluczowe: *eksploracja tekstów, klasyfikacja tekstów, język programowania R, analiza wydźwięku, leksykon wydźwięku polskich słów, uczenie maszynowe, Twitter*

## Sentiment analysis model for Twitter data in Polish language

Summary in English.

Text mining analysis of tweets gathered during Polish presidential election on May 10th, 2015. The project included implementation of engine to retrieve information from Twitter, building document corpora, corpora cleaning, and creating Term-Document Matrix. Each tweet from the text corpora was assigned a category based on its sentiment score. The score was calculated using the number of positive and/or negative emoticons and Polish words in each document. The result data set was used to train and test four machine learning classifiers, to select these providing most accurate automatic tweet classification results. The Naive Bayes and Maximum Entropy algorithms achieved the best accuracy of respectively 71.76% and 77.32%. All implementation tasks were completed using R programming language.

# 1. Introduction

## 1.1 Importance of mining the data on the internet

Today's social networks like Facebook, Twitter, Google+, LinkedIn and Weibo are populated by hundreds of millions of people each. These people often communicate, or comment on events in real time, what present real incentives, and at the same time new challenges for analytics and knowledge discovery in large datasets. Traditional Data Warehouse oriented Business Intelligence tools designed for structured data sources, are nowadays being supplemented by new tools and methods allowing effective social network analysis. As a result social media analytics have become the next steps in the evolution of traditional Business Intelligence. Companies go beyond traditional CRM systems or data warehousing, and begin having insight into client experience by following social networks. Monitoring, alerting and reporting on them, directing attention to brand or product can prove useful for several business applications. Moreover, data extracted from social networks like Twitter are increasingly being used to build applications and services to monitor and report on public reactions to events, political polarization, elections, protests, identification of epidemics, earning money in financial markets, or even the spread of misinformation. This revolutionizes not only the private sector and modern business environment, but also the whole public space, both on national and international levels.

In the 2011 Russian legislative election, Twitter was used intensively during protests tied to parliamentary elections as both pro-Kremlin and anti-Kremlin parties posted to Twitter to express their opinions. A recent study [1] of this election examined the mechanics of attacks launched by an unknown group that leveraged 25,860 accounts to send 440,793 tweets targeting 20 hashtags in protest of the election results. The authors estimate that at its peak a spam-as-a-service program controlled at least 975,283 Twitter accounts and mail.ru email addresses. Only 1% of the IP addresses used by attackers originated in Russia. Despite the large volume of malicious tweets, Twitter's search relevance algorithm which weights popular content, eliminated 53% of the tweets sent during the attack compared to the real-time search results. This indicates a positive outlook with regards to future censorship attacks.

Another study [2] noted that the number of Twitter followers for a presidential candidate in the 2012 United States Presidential Election surged by over 110 thousand within one single day. Analysis showed that most of these followers are unlikely to be real people. With an increasing number of messages posted each second in various languages, the task of social network monitoring, and sentiment analysis requires new tools to deliver psychographic insights key to understanding the new audience of social network users.

Locally, a few microblogging sites were created in Poland after the initial success of Twitter in 2006. These were for example Blip, Spinacz.pl, Flaker, and Pinger. As of April 25, 2015 Pinger[1] is the only one with an active website. It advertises to have 507,000 users, 15,087,987 posts, 20,547,710 photos, 75,464 audio files and 133,340 video files. Polish 'Pinger

---

[1] http://www.pinger.pl/

- świetni ludzie' translates into 'Pinger - great people'. The microblogging service displays several photos of models and young people. Their content is dominated by graphical elements with limited textual information. Therefore, it is not an optimal source of information for opinion mining experiments. In this paper, the author will focus on a problem of automating a sentiment analysis for Twitter. It offers a free of charge public API to all its registered users. Additionally, in Poland all major politicians have registered their profiles on Twitter. Some of them exchange messages publicly with their foreign colleagues, or sometimes provoke controversy[2].

In author's opinion one of the subjectively interesting topics discussed on Twitter in Polish language at the time when this project was being prepared in April and May 2015 was the election of President of the Republic of Poland. All but one candidate had at least one active twitter account[3] during electoral campaign.

## 1.2 The problem of sentiment analysis on Twitter

Opinion mining or sentiment analysis have generated great interest recently due to its potential benefits in trend analysis. Recent research [6] indicates that automated sentiment analysis of natural language is challenging. It requires a deep understanding of explicit and implicit, regular and irregular, and syntactical and semantic language rules. Researchers interested in opinion mining face challenges of NLP's unresolved problems:
- coreference resolution, negation handling,
- anaphora resolution,
- named-entity recognition,
- word-sense disambiguation.

Sentiment analysis research is gradually distinguishing itself as a new and separate field, falling in between NLP and NLU. Unlike standard syntactical NLP tasks, such as summarization and automated categorization, opinion mining mainly focuses on semantic inferences and affective information associated with natural language, and doesn't require a deep understanding of text. The authors [6][7] envision sentiment analysis research moving toward content-, concept-, and context-based analysis of natural language text, supported by time-efficient parsing techniques suitable for big social data analysis. This is because microblogging websites like Twitter have evolved to become a source of varied information that can be analysed as people post real time messages about their opinions on a variety of topics, discuss current issues, complain, and express positive sentiment for products they use.

Sentiment analysis of documents collected from Twitter is a challenging task. Tweets are limited to 140 characters encouraging Twitter users to use varied emoticons and word shortening techniques. This has resulted in the emergence of new, internet specific type of text that uses a different and more informal vocabulary. Another challenge involves the ambiguity of many posts as well as skew and bias in training data [5]. These problems are being addressed by classifying documents two-dimensionally, using two distinct classifiers. One to categorize a document as polar or neutral, and the other to highlight positive or negative sentiment. There have been several attempts to approach sentiment classification problems, and it's still a field of active research [10][15].

---

[2] http://blogs.wsj.com/emergingeurope/tag/radoslaw-sikorski/
[3] List of twitter accounts in Appendix 6.4.1 'Presidential Candidates in the 1st Round'

## 1.3 Goal and scope

In this project, the author will focus on the task of text classification. More specifically, he will implement a sentiment analysis model using Machine Learning techniques, that will be applied to a dataset collected from Twitter to determine whether tweets are positive or negative. Twitter is an good source of large volumes of unstructured data on virtually every topic. But the content is subject to various types of noise, and poses challenges mentioned briefly in the previous chapter. From the data sourcing perspective this analysis will focus on short messages issued in relation to Polish Presidential Election on May 10th, 2015. The goal of this project is to implement an application that supports the sentiment analysis of Twitter data in the Polish language to monitor sentiments related to electoral campaigns.

To achieve that the author will use a sentiment scoring algorithm that requires lexicons of Polish words to classify tweets into positive or negative. The result dataset will be used to build, train and evaluate classifiers with the intention to perform automatic text classification via supervised learning techniques, without the requirement to run the sentiment scoring algorithm again. Based on the recent research results on the subject of classification [9][15], it was decided to evaluate Naive Bayes approach, Maximum Entropy, Support Vector Machines, and Tree classifiers to determine their suitability for this application. All implementation work with regards to this project was performed using the R programming language, leveraging on the multiple libraries it offers for text mining and machine learning.

This paper is organized into five chapters. Chapter 1 introduces the concept of data mining on Twitter and relates it to the goals of the project. Chapter 2 defines sentiment analysis, its areas of application, and highlights key groups of algorithms used for sentiment classification. Chapter 3 covers various aspects of implementation, whereas Chapter 4 focuses on the experiments conducted on the data collected from Twitter to create, train and evaluate text classification using four different algorithms. Chapters 5 Provides conclusions and highlights ideas for further research. Appendices section contains bibliography, list of abbreviations, some additional information about 2015 Polish Presidential Election on Twitter, and solutions to technical problems encountered during the development of this project.

# 2. Sentiment Analysis

## 2.1 Definition and types of sentiment analysis

People acquire language understanding skills through intensive learning during the first years of their lives. The learning continues throughout their lifetime. It is an iterative process that consist of two key elements of language acquisition: building lexicon, and relating entries in the lexicon through a lexical network and/or ontology [3]. The very same approach is adopted by NLP, a field of computer science and computational linguistics that focuses on interactions between computers and human (natural) languages. Sentiment analysis or opinion mining in the context of NLP is the application of text mining or computational linguistic methods to identify and extract subjective information from a data source. A fundamental task in sentiment analysis is classifying the polarity of a given text at the document, sentence, or feature/aspect level. There are different types of sentiment analysis possible:
- classification of a document expresses a positive or negative opinion,
- assigning a grade to a document, for example in the scale from 1 to 10 points,
- emotion focused, to understand what emotions the author expressed about the subject(s) of his opinion. This is actively being researched by affective computing, a new field of computer science working on systems that are capable of recognizing, interpreting and processing emotions[4],
- focusing on the elements of opinions targeting selected features of a subject. In case of a known category of a subject, this can be a well-defined set of features. Otherwise, if it is a new domain of research, it is possible to consider an automatic selection of parameters,
- focusing on the analysis of text describing two entities (or products). These methods, that identify named entities and references to them, are known as Named-entity recognition (NER). Research on such systems has been structured as taking an unannotated text and producing an annotated block of text that highlights the names[5],
- discovery and interpretation of sentences describing or comparing opinions on two entities (products or services), and focusing on opinionated text [4],
- classification of documents to a (subjectively) interesting topic to asses if a document is interesting, or to follow all publications for the area of interest,
- detecting opinions which are spam or are not relevant.

The most fundamental sentiment analysis is a problem of document classification into positive and negative opinions. This implies binary classification. Depending on the training data it is also frequently possible to analyse the opposite qualities related to:
- authors emotions e.g. angry or calm, happy or unhappy, sad
- emotions that the text is intended to convey to the reader.

In cases of such binary classification, it is often practical to work with three classes in order to increase the quality of classification. Apart from using only the classes of our interest, it is advisable to create an additional class for neutral examples. This technique allows the classifier to return more useful results that better reflect the nature of classified text [5]. The other technique to increase the quality of sentiment analysis examined is derived from examining relation between subjectivity detection and polarity classification. The research showed that previous subjectivity detection at the sentence level can compress reviews into much shorter extracts that still retain polarity information at a level comparable to that of the full review. Application of this approach can be especially useful in applying advanced sentiment analysis algorithms on a set of large text documents to increase the quality of classification and significantly reduce the processing time [5].

---

[4] http://affect.media.mit.edu/
[5] http://nlp.stanford.edu/software/CRF-NER.shtml

Sentiment analysis can be performed on different levels:
- at the document level - where we focus only on the overall sentiment expressed by the whole document,
- at sentence level - where we analyse sentiment of individual sentences, and based on that we calculate the sentiment score of a document,
- on components of sentences, sub-sentences - where we analyse the sentiment of sub sentences, and based on them calculate the sentiment of sentences, and later the whole document.

The lower the level of analysis, the more advanced method of consolidating scores have to be considered, and the more complex result computations will be. As an example, a sentiment of a sentence can be calculated as average score based on the words it contains. In such case, if a sentiment of sentences was pre-calculated, it would be possible to evaluate it against the results for individual sentence components like sub-sentences. In the sentence from Example 2.1.1 the average score of its terms would indicate either positive or neutral sentiment (there are positive terms like 'confident', 'well' 'prepared', 'positively'). Whereas if we additionally consider the sentiment score at the sentence component level, we could discover that the second sub-sentence (with just one negative term 'lost') drives an overall negative opinion about the subject.

**Example 2.1.1**
*The candidate looked confident, well prepared and was even very well received by the audience, but lost the debate after answering that question.*

## 2.2 Areas of application

Sentiment analysis algorithms are applied where the subject of analysis is text data, and the dataset is so large or complex that traditional manual data processing approaches are inadequate. The greatest known source of textual data is Internet.

The first common application of sentiment analysis methods is opinion mining on brands, products, and services. Such analysis allows automated monitoring of how Internet users, who are existing or prospective clients of a company, perceive this company, its brands, products or services. This includes research on brand value, effectiveness of marketing campaigns, measuring the acceptance of new products, or initiating client support in case any initial signals of dissatisfaction are discovered.

Another common application of such algorithms is to allow the aggregation of opinions, and summarizing them in a form that is most relevant to their recipients. An example of that would be a calculation of average score for a product or group of products, or balancing the number of positive and negative opinions displayed to a user of an e-commerce website. Automated calculation of scoring can also be performed on a more granular level of every feature. In this way every feature of every product can be summarized, and the user is able to filter on the most relevant or interesting summary.

Yet another popular application of sentiment analysis is to measure and understand public opinions on events[6], social or political campaigns or key challenges municipalities have with the services they offer[7]. An important application in democratic countries where the majority of population are active Internet users is to monitor and understand political support patterns and election processes. In the US in 2008[8] election campaigns introduced Internet strategies and very aggressive ways to reach out to voters basically saying that the Internet matters as much as those other communications channels. For politicians or their political cabinets, it might prove useful to try to understand voters, try to appeal to voters, try to get people to talk to their friends about the campaigns, or simply measure how they are being perceived by the wider public. It might also be useful to identify the key issues that are being discussed online, so that a political party or candidate can try address them. In recent years, social media has seen the development of new forms of spam relating to politics [6]. As manipulation of social media affects perceptions of candidates and compromises decision making, such techniques can manipulate the emotions of a society [1]. Additionally, the use or new abuse techniques, like "google bombs", "twitter bombs" or targeted Twitter spamming can apply pressure to modulate views and contribute to political censorship [6].

## 2.3 Types of algorithms for sentiment classification

One of the approaches to describe the evolution of sentiment analysis research attempts to view it from the perspective of analytical tokens, or building blocks, and the implicit information associated with those tokens. It groups the existing algorithms into four main types: keyword spotting, lexical affinity, statistical methods, and concept-based techniques [6]. These methods will be summarized in the remaining part of this chapter.

### 2.3.1 Keyword spotting

This is the most naive approach, and a frequently used method. Its popularity is driven by the ease and speed of use. This approach classifies text by desired categories based on the presence of popular affect words from a predefined dictionary. Such dictionaries for the English language are widely available. Keyword spotting has two main weaknesses. It can't reliably recognize affect-negated words. It is unable to classify sentences that do not contain any affect-terms from the dictionary, but convey obvious sentiment for humans through underlying meaning rather than affect adjectives. Although keyword spotting could correctly classify simple sentences containing known affect words, the text from the example below would unlikely be classified correctly as generating a strong sentiment.

---

[6] http://sc1.qcri.org/cop18/
[7] http://www-03.ibm.com/press/us/en/pressrelease/38816.wss
[8] http://fpc.state.gov/193458.htm

**Example 2.3.1**
*He recorded white Russian trucks crossing the border and progressing fast towards Białystok.*

The keyword spotting methods can be enhanced by:
- defining a strength of sentiments for affect words, in addition to their binary classification into positives or negatives in the lexicon,
- defining the whole expressions containing a number of words as positives or negatives in the lexicon,
- defining an additional category of words that modify (increase or decrease) the sentiment of positive or negative words or expressions from the lexicon.

## 2.3.2 Analysis of lexical affinity

Lexical affinity is slightly more sophisticated than keyword spotting. It does not only examine the frequency of affect words, but also assigns them a probable "affinity" to particular emotions. For example, lexical affinity might assign the word "war" a 90-percent
probability of indicating a negative sentiment. This approach usually trains probability from linguistic corpora. As the research show [6] although it often outperforms keyword spotting techniques, there are two main weaknesses in this approach:
- negated sentences and sentences with other meanings trick lexical affinity, because they operate solely on the word level,
- lexical affinity probabilities are often biased towards a particular type of text indicated by linguistic corpora. So, it is difficult to develop a re-usable, domain-independent model.

**Example 2.3.2**
*I avoided the war.*
*Price war continued in Q3.*

## 2.3.3 Statistical methods

This approach, which is popular for affect text classification, includes four popular algorithms: Naive Bayes, Stochastic gradient descent, Support vector machines, and Maximum entropy. Researchers apply statistical methods by providing a large training corpus of affectively annotated texts as input to machine-learning algorithms [9]. The system's classifier might not only learn the affective valence of affect keywords (as in the keyword spotting approach), but also take into account the valence of other arbitrary keywords (similar to lexical affinity), punctuation, and word co-occurrence frequencies [6]. Statistical methods are considered to be semantically weak, which means that co-occurrence elements have little predictive value. Moreover such text classifiers only work well when they receive sufficiently large text input. As a result, even if these methods might be able to effectively classify a user's text on the page level or paragraph level, they will not necessarily work well on sentences or components of sentences. The other potential weakness is that because the amount of data to be analysed is usually large, they could take a significant amount of time to execute which can be impractical in some applications. Of the four mentioned algorithms described as statistical methods, the best performing algorithm applied on Twitter data that was able to achieve accuracy, and process significantly larger training sets in a reasonable amount of time was Naive Bayes. The researchers [9] were able to train the Naive Bayes classifier on 1.6 million Tweets, which gave the algorithm almost 80% accuracy, outperforming the other algorithms.

### 2.3.4 Concept-based techniques

Opinion mining models using concept-based techniques use semantic networks to accomplish semantic text analysis [6]. They rely on large semantic knowledge bases. In this way they allow a system to grasp the conceptual and affective information associated with expressing different sentiments in natural language opinions. Such approaches focus on implicit features associated with natural language concepts and can analyse multi-word expressions that don't explicitly convey emotion, but are related to concepts that do.

The concept-based approach relies heavily the size and quality of knowledge bases it uses, and requires a definition of grammar rules to build ontologies or semantic networks. A comprehensive, universal, textual databases that encompasses up-to-date human knowledge exist and can be used for knowledge discovery. These sources however are often inappropriate for sentiment analysis, as they tend to use formal language that does not include slang, jargon or contain nuances that are atypical. Opinion mining systems processing the semantics of natural language text often encounter other challenges, like these around the need to define the grammar rules that are specific for natural languages. Such rules can be complex or ambiguous for many of the modern natural languages.

# 3. Implementation

This chapter describes the technologies that were used in this project, key components of implemented sentiment analysis application, as well as training and testing datasets used to produce results described in Chapter 4 of this paper. This implementation was inspired by two articles [13][16].

## 3.1 Technologies

All the tools supporting this project were implemented using the R programming language. This language was chosen because it offers high level of abstraction and simplicity in operating on various data structures. It also offers a wealth of libraries delivering multiple statistical, graphical, and machine learning capabilities out-of-the box. Moreover, it integrates with a convenient IDE. To implement a custom twitter sentiment analysis tool the author used the following open-source tools:

- R version 3.2.0 (2015-04-16) running on i686-pc-linux-gnu[9],
- RStudio version 0.98.1103[10],

and libraries:
- twitteR - R based Twitter client that provides an interface to the Twitter web API[11],
- dplyr - R package for working with data frames both in memory and out of memory,
- stringi - R package for string/text processing in each locale and any native character encoding[12]. It was developed by Marek Gagolewski and Bartłomiej Tartanus who graduated from Warsaw University of Technology,
- ggplot2 - R package used for graphical presentation of data[13],
- tm - R package for text mining applications within R[14],
- e1071 - R package used to create a Naive Bayes classifier, developed by various authors related to the Department of Statistics, TU Wien[15],
- RTextTools - R package used to enhance the functionality of e1071 in relation to automatic text classification via supervised learning with multiple additional learning algorithms ("MAXENT", "SVM", "RF", "BAGGING", "TREE").

The author provides more details on the configuration of the development workstation with regards to R in the appendices section (Chapter 6.3.1). The hardware was changed during the execution of experiments with classifiers (Chapters 6.3.1 and 6.3.5).

## 3.2 Acquisition of test and train data

As mentioned in the goals section, the author decided to focus sentiment analysis from data perspective on short messages issued on Twitter in relation to the electoral campaign before the Polish Presidential Election on May 10th, 2015. As this event was not subject to any publicly available text corpora, He decided to create one by collecting tweets by candidates and their followers. His high level approach included the following steps:

1. Confirming the official list of all people who registered themselves as candidates in the 2015 Polish Presidential Election. Such a list can be found on the website of National Electoral Committee[16]
2. Manually searching Twitter for profiles of people who identified themselves as candidates in the Polish Presidential Election. He recorded all the active accounts in the appendices section (Chapter 6.4.1) on April 12, 2015 and discovered that only one candidate (Grzegorz Michał Braun) was not an active user of Twitter. If candidates had several active profiles, he compared them to identify the single one that was established in relation to presidential campaign
3. Collecting tweets from each candidate and saving them to disk. In order to achieve that he used an R script called scraper.R which connects to the Twitter API using the author's credentials, and requests 1500 tweets related to each candidate's profile (from step 2) using the *searchTwitter()* function from the twitteR package. Excluding from consideration the limits of Twitter API, the first challenge he encountered was that not all searches produced 1500 tweets, despite several attempts to collect an equal number of tweets for each candidate profile. The author realised that a few candidates did not produce 1500 messages in total on Twitter, and noticed that some profiles are clearly more active and popular than others. For these less active profiles, as he collected significantly more messages in the dataset then each of these profiles posted on Twitter (E.g. @JacekWilkPL - 434 tweets, @Pawel_Tanajno - 459 tweets, and @M_Kowalski1 - 1189 tweets). The author assumed that the collected number of tweets will provide a representative sample adequate for the sentiment classification task.
4. Most experiments described in this paper use the subset of collected tweets. The subset of 6040 tweets tagged as positive or negative, excluding neutral tweets.
5. The data have been divided into training set and test set using the 70/30 principle. 70% of randomly selected records have been selected into the training set, and the remaining 30% become the test set.

---

[9] http://www.r-project.org/
[10] http://www.rstudio.com/
[11] http://cran.r-project.org/web/packages/twitteR/index.html
[12] http://www.rexamine.com/resources/stringi/
[13] http://ggplot2.org/
[14] http://cran.r-project.org/web/packages/tm/index.html
[15] http://cran.r-project.org/web/packages/e1071/index.html
[16] http://prezydent2015.pkw.gov.pl/306_Kandydaci

**Exhibit 3.2.1 Preparation of test and train data for experiments**

```
> set.seed(1234)
> TtTs <- sample(2, nrow(raw.corpus), replace=TRUE, prob=c(0.7, 0.3))
> trainData <- raw.corpus[TtTs==1,]
> nrow(trainData)
[1] 4255
> testData <- raw.corpus[TtTs==2,]
> nrow(testData)
[1] 1785
```

# 3.3 Sentiment analysis tool implemented with R

The author intended to reuse as many elements of the R framework (presented by Jeffrey Breen during the Boston Predictive Analytics MeetUp in 2012 [13]) as possible. However, due to different goals of this project he had to introduce several changes to what he proposed. The code of the project is organized into the following folders:
- 'data', that contains:
    - 10 input data files (tweets) in .RData format that were collected from Twitter for each candidate using scrapper.R script (described in Chapter 3.2, step 3)
    - sub-folder 'opinion-lexicon-Polish' with 2 text files positive-words.txt, negative-words.txt that are the 2 opinion lexicons described in Chapter 3.4.1
- 'output', that is intended for data visualizations generated with ggplot2 in .pdf format, and any other output data files.
- 'R', that contains several R scripts:

- main.R - the main script, installing required packages and useful for running the other scripts
- scrapper.R -  collecting tweets and caching them to a directory
- functions.R - containing all custom functions
- 0_start.R - set up environment variables, specify path locations, load prerequisite libraries and all custom functions like score.sentiment()
- 1_load.R - loads Twitter data and sentiment lexicon
- 2_preprocess.R - processes the tweets and visualizes the dataset. This is to prepare the data for Machine Learning part; scraper.R should be run once to collect and cache tweets before running this script.
- 3_ml.R - Machine Learning tasks, building, training and evaluating Naive Bayes classifier and other methods.

### 3.3.1 Function for calculating sentiment score

There are potentially many ways to calculate sentiment scores for opinion mining and sentiment analysis. This project uses a simple approach to define a score formula [13] suggested by Jeffrey Breen. His general idea was to calculate a sentiment score for each tweet and use that score to evaluate whether a positive or negative message was posted. He proposed using a bag-of-words representation of a message to calculate its sentiment score using the simple formula described below. When this paper relates to a positive document (or tweet) this means that the sentiment score value for a given document is greater than zero (0 being neutral). If the value is less than 0, then such document is considered as negative.

**Exhibit 3.3.1 Sentiment score calculation formula**

```
SentiScore  =  Number of positive words  -  Number of negative words

If Score > 0, this means that the sentence has an overall 'positive opinion'
If Score < 0, this means that the sentence has an overall 'negative opinion'
If Score = 0, then the sentence is considered to be a 'neutral opinion'
```

In order to count the number of positive and negative words the author reused an opinion lexicons for the English language provided by Hu and Liu[17]. The two (positive and negative) lexicons were not well placed for the task of annotating the corpus of tweets written by presidential candidates in the Polish language (as defined in Chapter 1.3). Therefore, the author decided to create a similar lexicons for the Polish language of positive and negative words. Additionally, he decided to supplement the original approach by the inclusion of positive and negative emoticons that are frequently used on twitter to express emotions. These lexicons have been described in the following chapter.

Another important element shared by Jeff Breen, is the function to calculate sentiment score. He originally wanted to reuse this function [13] to perform the analysis. Unfortunately it was not working with the Polish tweets that he collected, so it had to be rewritten it into two functions *prepare.for.sentiment <- function(sentences)* and *score.sentiment <- function(sentences, pos.words, neg.words)*. The key changes and optimizations proposed were:

- Use of R package dplyr instead of plyr. Dplyr is the next iteration of plyr. It is expected to be faster in data processing and has a more consistent API[18].
- Use of R package stringi instead of stringr to allow text processing of Polish character encoding.
- Replacement of letters specific to the Polish alphabet using a UTF-8 table[19] into basic latin characters to simplify the processing of Polish words using the bag-of-words representation [13]. This was to avoid duplicating words in positive and negative lexicons to include/exclude Polish letters in words in different (random) positions in a word. The author believed it is worth assuming that more informal texts, like micro-blog posts in the Polish language are more likely to be written incorrectly with the Latin alphabet only, or to contain other errors like using Polish characters on a few, selected positions only.
- Use of emoticons as additional vehicles contributing to overall positive or negative sentiment score. The function replaces negative emoticons with word 'neg.emot' and positive emoticons with word 'pos.emot'. Both words were added to relevant positive and negative lexicons described in the following chapter.

---

[17] http://www.cs.uic.edu/~liub/FBS/sentiment-analysis.html

**Exhibit 3.3.2 Custom sentiment scoring functions**

```
prepare.for.sentiment <- function(sentences)
{
        require(dplyr)
        require(stringi)

        sentences %>%
        stri_trans_tolower() %>%
        stri_replace_all_fixed("\u0105", "a") %>%
        …
        stri_replace_all_fixed("\u017C", "z") %>%

        stri_replace_all_fixed(":)", "pos.emot") %>%
        …
        stri_replace_all_fixed("=D", "pos.emot") %>%
        …
        stri_replace_all_fixed(":(", "neg.emot") %>%
        …
        stri_replace_all_fixed("8C", "neg.emot") -> prepared

        return(prepared)
}
score.sentiment <- function(sentences, pos.words, neg.words)
{
        scores <- lapply(sentences, function(sentence, pos.words, neg.words)
        {
        require(dplyr)
        require(stringi)

        sentence %>%
        stri_extract_all_words() %>%
        unlist() %>%
```

---

[18] http://blog.rstudio.org/2014/01/17/introducing-dplyr/
[19] http://utf8-chartable.de/unicode-utf8-table.pl

```
        na.omit() -> words

        pos.matches <- sum(words %in% pos.words)
        neg.matches <- sum(words %in% neg.words)

        score <- pos.matches - neg.matches
        return(score)

    }, pos.words, neg.words )

    scores.df <- data.frame(SentiScore=scores, Tweet=sentences)
    return(scores.df)
}
```

### 3.3.2 Lexicon of positive and negative words

The lists of positive and negative words have been created manually, originally by translating the two sentiment lexicons for the English language prepared by Bing Liu[20]. Additionally, the author has added additional Polish words that he subjectively considered positive or negative, what included words from texts that were collected from Twitter. The sentiment lexicon was organized into two text files containing two exclusive lists sorted alphabetically and expressed in the standard Latin alphabet (disregarding additional letters in the Polish alphabet):

- 2000 Polish words classified as positive,
- 3693 Polish words classified as negative.

Each list includes one special word: *pos.emot* and *neg.emot*. They are considered placeholders for 18 positive and 22 negative emoticons used by *prepare.for.sentiment()* (Chapter 3.3.1). Emoticons are popular forms of expressing emotions, especially in short messages or informal communication. They are also popular on twitter and easily classifiable as positive or negative.

### 3.3.4 Classification algorithms

An algorithm that implements classification is known as a classifier. The term 'classifier' also refers to a mathematical function, implemented by a classification algorithm, that maps input data to a category. Bayesian methods have been used in pattern recognition in various studies 20 years prior to being adopted by machine learning researchers in the early 1990s to classify redundant attributes and later numeric attributes. As the authors [11] suggest, the term 'naive' is unfortunate because the approach is intuitive and practical for large data inputs. If used in the appropriate circumstances, the Naive Bayes model is particularly appropriate for text classification (McCallum and Nigam, 1998). The Bayesian Method described in [11] has been added to this project by incorporating functionality from the e1071 R package.

**Exhibit 3.5.1 naiveBayes() function**

```
mat = as.matrix(matrix.sparse)
classifier = naiveBayes(mat[TtTs==1,], as.factor(raw.corpus[TtTs==1,4]) )
predicted = predict(classifier, mat[TtTs==2,]);
```

---

[20] http://www.cs.uic.edu/~liub/FBS/sentiment-analysis.html#lexicon

The other classifiers, Naive Bayes, Maximum Entropy, Support Vector Machines, Tree have been implemented using functions from RTextTools R package.

**Exhibit 3.5.2 Maximum Entropy, Support Vector Machines, and Tree functionality**

```
container = create_container(matrix, as.numeric(as.factor(raw.corpus2[,4])),
                trainSize=1:4255, testSize=4256:6040,virgin=FALSE)
models = train_models(container, algorithms=c("MAXENT","SVM", "TREE"))
predicted = classify_models(container, models)
```

# 4. Experiment Evaluation

## 4.1 Dataset

The raw dataset collected for this project was downloaded from Twitter using its REST API described in the appendices section (Chapter 6.3.4). It contains 11,744 tweets related to 10 profiles of candidates in the 2015 Polish presidential election (see the appendices section, Chapter 6.4.1 for a full list of twitter accounts). The tweets. which were from May 10th, 2015 between 1.10-1.30 AM CET, annotated using sentiment score calculation from Chapter 3.1.1 and visualized with the tool implemented in R for this project (Chapter 3.3). The sentiment score values for different tweets in the dataset range from -5 to 6. The distribution of positive and negative score resembles a bell-shape, with a slightly left-skewed distribution. A minor asymmetry towards positive scores could be attributed to the natural tendency of presidential candidates (politicians) to deliver a positive message in electoral campaigns. Considering the naive sentiment scoring approach used in this project and the limited dataset, there was only one candidate in the 2015 Polish Presidential Elections who could be described as having a negative campaign on Twitter (number 6).

**Exhibit 4.1.1 Distribution of positive and negative sentiment scores per candidate in the text corpora**

```
> all.scores %>% filter(SentiScore != 0) %>%
+   select(Candidate, SentiScore) %>% group_by(Candidate) %>%
+   summarise( Tweets = n(), Median = median(SentiScore), Mean = mean(SentiScore),
+             StdDev = sd(SentiScore), Min = min(SentiScore), Max = max(SentiScore))
Source: local data frame [10 x 7]

                   Candidate Tweets Median        Mean   StdDev Min Max
1       Duda Andrzej Sebastian    746      1  0.82975871 1.109676  -2   5
2       Jarubas Adam Sebastian    351      1  0.62962963 1.239645  -3   3
3   Komorowski Bronisław Maria    727      1  0.18569464 1.343247  -4   5
4  Korwin-Mikke Janusz Ryszard    664      1  0.40512048 1.314740  -4   4
5      Kowalski Marian Janusz    737      1  0.81818182 1.216751  -3   4
6           Kukiz Paweł Piotr    836     -1 -0.05023923 1.298497  -3   4
7   Ogórek Magdalena Agnieszka    742      1  0.55929919 1.351208  -3   5
8       Palikot Janusz Marian    737      1  0.46540027 1.298707  -5   6
9            Tanajno Paweł Jan    256      1  0.14062500 1.238179  -3   3
10                 Wilk Jacek    244      1  0.56557377 1.220455  -2   4
```

**Exhibit 4.1.2 Histograms presenting all sentiment scores calculated on the text corpora - per candidate and in total.**

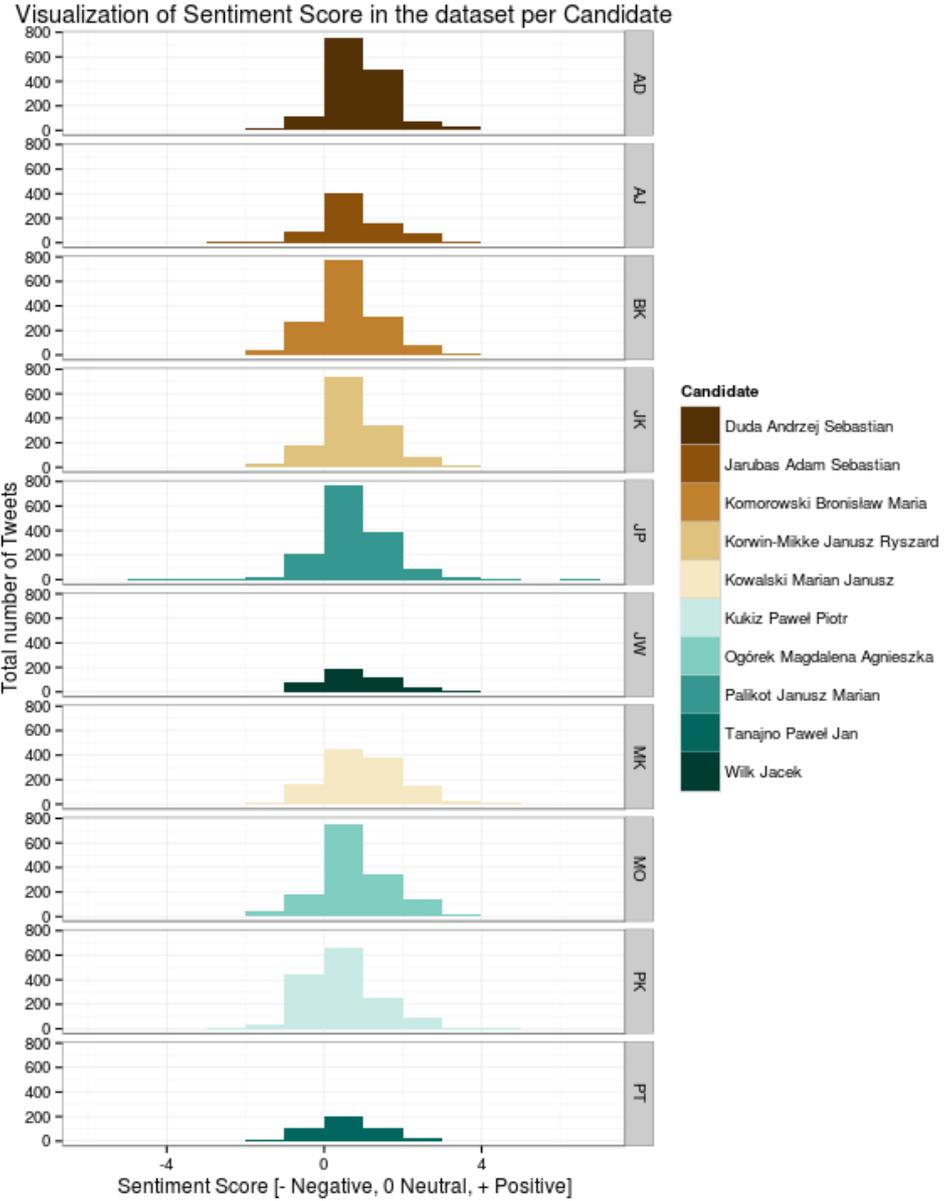

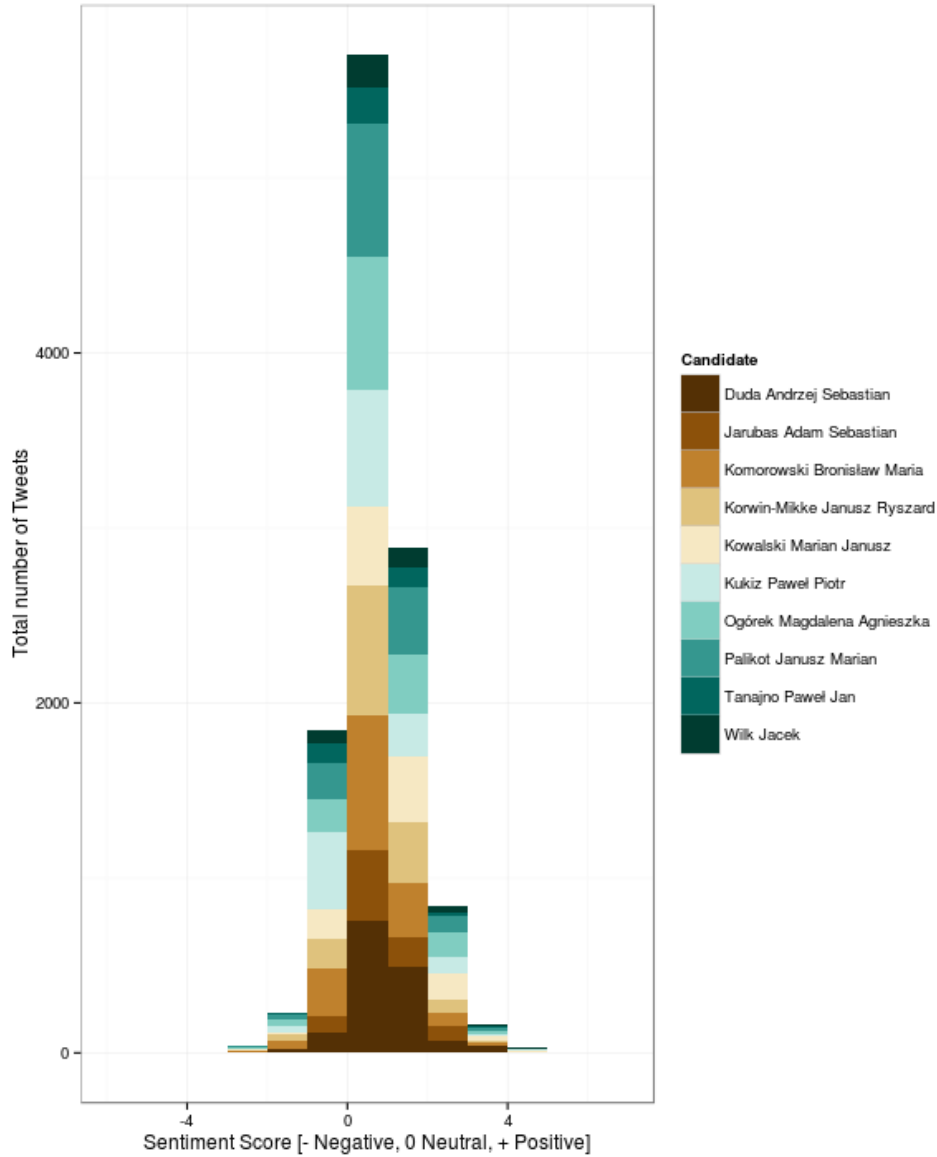

The textual dataset containing 11,744 tweets related to 10 presidential candidates have been used to evaluate Naive Bayes, and various other machine learning algorithms for their usefulness with regards to sentiment classification of tweets into two or three classes:
- positive / negative,
- positive / neutral / negative.

**Exhibit 4.1.3 Summary of the full dataset, including neutral tweets.**

```
> summary(raw.corpus)
   SentiScore                                                          Tweet            Code              Candidate          SentiClass
 Min.   :-5.0000   rt @prezydentkukiz: @komorowski tak,                                                                                
 1st Qu.: 0.0000   rt @m_kowalski1: co do konfliktu na l  e jow... teraz juz wiem - to byla...:  162   Length:11744     Length:11744     Length:11744    
 Median : 0.0000   rt @tadeuszsznuk: .@ogorekmagda pani   mi swoje glosy pos.emot            :  103   Class :character  Class :character  Class :character
 Mean   : 0.2331   rt @andrzejduda2015: tymczasem na #du                                      :   93   Mode  :character  Mode  :character  Mode  :character
 3rd Qu.: 1.0000   rt @katarzynapawlak: zaorane.\n@prezy  stifmt                              :   85                                                     
 Max.   : 6.0000   rt @tadeuszsznuk: .@m_kowalski1 panie                                      :   84                                                     
                   (other)                                mozna pomylic sie o 5 kg.          :   84                                                     
                                                                                             :11133                                                     
```

```
> str(raw.corpus)
'data.frame':  11744 obs. of  5 variables:
 $ SentiScore: int  0 -1 0 0 1 0 2 0 0 -2 ...
 $ Tweet     : Factor w/ 7939 levels "10 maja zaglosuje: na wolnosc, na dumna i bogata polske,
 $ Code      : chr  "JK" "JK" "JK" "JK" ...
 $ Candidate : chr  "Korwin-Mikke Janusz Ryszard" "Korwin-Mikke Janusz Ryszard" "Korwin-Mikke
 $ SentiClass: chr  "Neutral" "Negative" "Neutral" "Neutral" ...
```

**Exhibit 4.1.4 Summary of the dataset excluding neutral tweets**

```
> summary(raw.corpus2)
  Candidate                                     Tweet          SentiScore       SentiClass
 Length:6040        rt @prezydentkukiz: @komorow uz wiem - to byla...: 162   Min.   :-5.0000   Length:6040
 Class :character   rt @m_kowalski1: co do konfl os.emot           : 103   1st Qu.:-1.0000   Class :character
 Mode  :character   rt @andrzejduda2015: tymczas                   :  85   Median : 1.0000   Mode  :character
                    rt @tadeuszsznuk: .@m_kowals ie o 5 kg.         :  84   Mean   : 0.4531
                    rt @prezydentkukiz: ja nie w /2                 :  72   3rd Qu.: 1.0000
                    rt @partia_korwin: tu zestaw tp://t.co/hqnx74b...:  64   Max.   : 6.0000
                    (Other)                                        :5470
```

     The author encountered a few challenges when using 'tm' package to create the corpora and explore its contents. The attempts to process the entire dataset containing tweets related to all 10 candidates were initially not successful. He overcame this limitation gradually reducing the number of tweets processed from 11 thousand to 1468. This was the first break even number of tweets that he was able to process using tm package[21]. This unplanned "pre-experiment" resulted in changing author's development workstation to a different computer. The problem and remediation approach that was used is described in the appendices (Chapter 6.3.5). Despite these challenges, the words from each tweet tagged as positive, neutral or negative were used to train several types of classifiers, and evaluate their accuracy with regards to the automatic classification of new tweets.

## 4.2 Sentiment classifiers and their evaluation metrics

     The classifiers evaluated as part of this project have been build using several different algorithms: Naive Bayes, Maximum Entropy, Support Vector Machines, and Tree. The key evaluation metric the author considered in this paper is *accuracy* of classifier. The other is *recall*, also known as sensitivity or TPR. It refers to the proportion of cases in a class that the algorithm correctly assigns to that class. The other support metric that the author considered using to achieve very similar results for accuracy, and/or recall accuracy was *precision* or PPV. This refers to how often a case that the algorithm predicts as belonging to a class actually belongs to that class. The last metric that could additionally be considered is *F-scores*. It represents a weighted average of both precision and recall, where the highest level of performance is equal to 1 and the lowest 0. All of these metrics can be summarized using *create_analytics()* function from the RTextTools package which produces precision, recall and f-scores for analysing algorithmic performance at the aggregate level [17].

---

[21] The unit test was to calculate sum of frequency of terms in the Document Term Matrix for the text corpora of this size

**Exhibit 4.2.1 Selected formulas for calculating Accuracy, Recall and Precision**

*Accuracy = Σ True Positive + Σ True Negative / Σ Total Population*
*Recall = Σ True Positive / Σ Condition Positive*
*Precision = Σ True Positive / Σ True Positive + Σ False Positive*

## 4.3 Experiment results

**Exhibit 4.3.1 Experiment result summary with regards to the Naive Bayes classifier**

| No | Brief experiment description | Confusion matrix | | | Accuracy *) Precision |
|---|---|---|---|---|---|
| 0 | Initial processing of 1000 tweets. The algorithm runs. Execution time: a few seconds. | | Negative | Positive | 32.12% (0.3211921) |
| | | Negative | 97 | 5 | |
| | | Positive | 200 | 0 | |
| 1.1 | First attempt to process two classes: positive and negative tweets only. Text corpora of 6040 tweets:<br>● 3920 tagged as positive,<br>● 2120 tagged as negative.<br>Processing time: 5 minutes. | | Negative | Positive | 35.39% (0.3539326) |
| | | Negative | 630 | 0 | |
| | | Positive | 1150 | 0 | |
| 1.2 | Experiment 1.1 run with DTM using two flags (removePunctuation = TRUE, stripWhitespace = TRUE). | | Negative | Positive | 35.39% (0.3539326) |
| | | Negative | 630 | 0 | |
| | | Positive | 1150 | 0 | |
| 2 | Experiment with three classes of tweets:<br>● 2120 tagged as negative,<br>● 5704 tagged as neutral,<br>● 3920 tagged as positive. | | Negative | Neutral | Positive | 18.10% (0.1809632) |
| | | Negative | 635 | 0 | 13 | |
| | | Neutral | 1693 | 0 | 4 | |
| | | Positive | 1164 | 0 | 0 | |
| 3 | Experiment run on two classes, on sparse terms only (removeSparseTerms(matrix, sparse = 0.99)). | | Negative | Positive | 63.87% (0.6386555) |
| | | Negative | 573 | 32 | |
| | | Positive | 613 | 567 | |
| 4 | Experiment run on two classes, on sparse terms only. | | | | 67.06% |

| | | | Negative | Positive | (0.6705882) |
|---|---|---|---|---|---|
| | (removeSparseTerms(matrix, sparse = 0.90)) | Negative | 551 | 54 | |
| | | Positive | 534 | 646 | |
| 5 | Experiment run on two classes, on sparse terms only. (removeSparseTerms(matrix, sparse = 0.95)) | | Negative | Positive | 71.76% (0.7176471) 80.84%*) |
| | | Negative | 395 | 210 | |
| | | Positive | 294 | 886 | |
| 6 | Experiment run on all words (no removeSparseTerms flag), two tweet classes: positive and negative only. | | Negative | Positive | 33.89% (0.3389356) |
| | | Negative | 605 | 0 | |
| | | Positive | 1180 | 0 | |
| 7 | Two classes, sparse terms only. (removeSparseTerms(matrix, sparse = 0.995)) | | Negative | Positive | 70.59% (0.7058824) 87.77%*) |
| | | Negative | 499 | 106 | |
| | | Positive | 419 | 761 | |

    It was confirmed that the Naive Bayes algorithm performs much better when trained on the Term-Document Matrix containing fewer terms and less noise. Classifiers trained on all words of the original tweet representation were useless for automatic classification of new tweets. Assuming such document representation it only delivered 34% of overall accuracy and did not manage to classify any true positive tweets correctly. If used to classify tweets to three classes, positive, neutral and negative it only achieves 18.1% accuracy. However, when trained on the matrix with a parameter sparse = 0.995 it provided almost 71% accuracy and high precision of 87.77%. Such results indicate that the Naive Bayes algorithm, when applied to classify new tweets, can assign a vast majority of them correctly.

    Other machine learning classifiers evaluated as a part of this project on the same dataset were Maximum Entropy, Support Vector Machines, and Tree algorithm.

**Exhibit 4.3.2 Experiment result summary with regards to Maximum Entropy, Support Vector Machines, and Tree classifiers**

| Algorithm name | Recall Accuracy | Precision *) | F-score *) |
| --- | --- | --- | --- |
| Maximum Entropy | 55.41% (0.5540616) | 0.515 | 0.515 |
| Support Vector Machines | 64.31% (0.6431373) | 0.545 | 0.400 |
| Tree | 35.51% (0.3551821) | 0.330 | 0.400 |

*) Calculated automatically by create_analytics() from RTextTools R package, from algorithm performance section.

**Exhibit 4.3.3 10-fold cross-validation results for Maximum Entropy, Support Vector Machines, and Tree classifiers**

```
> cross_validate(container,N,"SVM")
Fold 1 Out of Sample Accuracy = 0.6478632
Fold 2 Out of Sample Accuracy = 0.6433022
Fold 3 Out of Sample Accuracy = 0.6172414
Fold 4 Out of Sample Accuracy = 0.6547434
Fold 5 Out of Sample Accuracy = 0.6687598
Fold 6 Out of Sample Accuracy = 0.6588629
Fold 7 Out of Sample Accuracy = 0.6525424
Fold 8 Out of Sample Accuracy = 0.6597111
Fold 9 Out of Sample Accuracy = 0.6551127
Fold 10 Out of Sample Accuracy = 0.6424779
[[1]]
 [1] 0.6478632 0.6433022 0.6172414 0.6547434 0.6687598 0.6588629 0.6525424 0.6597111 0.6551127 0.6424779

$meanAccuracy
[1] 0.6500617

> cross_validate(container,N,"TREE")
Fold 1 Out of Sample Accuracy = 0.6239168
Fold 2 Out of Sample Accuracy = 0.6285211
Fold 3 Out of Sample Accuracy = 0.6525822
Fold 4 Out of Sample Accuracy = 0.6695502
Fold 5 Out of Sample Accuracy = 0.6805112
Fold 6 Out of Sample Accuracy = 0.6580227
Fold 7 Out of Sample Accuracy = 0.6460033
Fold 8 Out of Sample Accuracy = 0.6372549
Fold 9 Out of Sample Accuracy = 0.6172007
Fold 10 Out of Sample Accuracy = 0.6726094
[[1]]
 [1] 0.6239168 0.6285211 0.6525822 0.6695502 0.6805112 0.6580227 0.6460033 0.6372549 0.6172007 0.6726094

$meanAccuracy
[1] 0.6486172
> N=10
> set.seed(2015)
> cross_validate(container,N,"MAXENT")
Fold 1 Out of Sample Accuracy = 0.7678571
Fold 2 Out of Sample Accuracy = 0.8029316
Fold 3 Out of Sample Accuracy = 0.7692308
Fold 4 Out of Sample Accuracy = 0.7496206
Fold 5 Out of Sample Accuracy = 0.7621359
Fold 6 Out of Sample Accuracy = 0.770206
Fold 7 Out of Sample Accuracy = 0.7729549
Fold 8 Out of Sample Accuracy = 0.7819315
Fold 9 Out of Sample Accuracy = 0.7890071
Fold 10 Out of Sample Accuracy = 0.7657993
[[1]]
 [1] 0.7678571 0.8029316 0.7692308 0.7496206 0.7621359 0.7702060 0.7729549 0.7819315 0.7890071 0.7657993

$meanAccuracy
[1] 0.7731675
```

The overall summary of accuracies considered for all classifiers evaluated in this project on the dataset of 6040 tweets (positive/negative) is as follows:
- Naive Bayes 70.59%
- Maximum Entropy 77.32%
- Support Vector Machines 65%
- Tree 64 %

The Naive Bayes and Maximum Entropy algorithms achieved the best accuracy of respectively 71.76% and 77.32%, and they are expected to perform well with classification of new tweets.

# 5. Conclusions

## 5.1 Summary

The importance of mining textual data on the Internet continues to grow. Twitter becomes the important mean of communication. Opinion mining activities, even on a relatively small dataset requires careful planning and preparation. Data exploration and supervised machine learning on small datasets collected from a micro-blogging site like Twitter is possible with limited computing resources. Real world problems require intensive computer processing power and should be planned carefully in relation to the required capacity of computer resources, time and with optimization techniques in mind. There are various open-source tools, development frameworks, and libraries that support sentiment analysis and/or other types of data analysis tasks on the R platform. These tools are widely available, and can be used free of charge, but require preliminary knowledge, with support being offered online on by a relevant community. The quality of sentiment analysis depends vastly on the quality of the dictionaries used, and the application of non-trivial, natural language processing methods.

The project was successfully completed because it delivered both the required dataset and tools supporting sentiment analysis on Twitter in the Polish language. A simple sentiment scoring algorithm was used with the Polish language by creating lexicon of positive and negative Polish words. 6040 tweets were annotated using this algorithm, and the textual data from these documents was used to experiment with building, training and evaluating Naive Bayes, Maximum Entropy, Support Vector Machines, and Tree classifiers. Two approaches: Naive Bayes and Maximum Entropy achieved the best accuracy (71.76% and 77.32% respectively) and can be used to classify the sentiment of new tweets collectable through the Twitter API.

## 5.2 Further Research

Throughout late-April, May and June 2015, when the author was working on this project, he had a few additional ideas related to further research, and next steps in development of a sentiment analysis application. These ideas, mostly related to the application enhancement, were descoped from the initial release in this paper due to time constraints, and maintaining focus on the primary goal of this project. This goal was to implement and evaluate a sentiment analysis model, an algorithm allowing automated classification of new tweets into positives and negatives. Nevertheless, there is a plan to explore at least a few of the below mentioned ideas in future.

The first idea is to incorporate additional sources of textual information available on the (Polish) Internet. The priority should be on including Facebook posts and the 'like' feature. Many candidates were present on Facebook, and their opinions were not included, nor even evaluated for use in the text corpora. This additional data from Facebook could be acquired with relatively limited effort, as the custom tool (described in Chapter 3.3) could connect to the Facebook API using predefined functions from the R package called Rfacebook[22]. Other popular Polish Social networking sites like nk.pl or pinger.pl, could be evaluated for potential benefits in balancing the text corpora used by the application. The author believes that the same evaluation could be performed for leading local news sites and web portals popular in Poland, like Google News[23], Interia[24], Onet[25] or Wirtualna Polska[26]. If proven useful, a development project to include additional web scraping code could be facilitated by using R package called XML (described in Chapter 6.3.3). This could be more challenging because each portal structures its information in a different way, and currently there is no universal API for all of them. As a result, the tool could evolve towards a general application capable of monitoring political activity and providing a sentiment index for candidates who are present online.

Commercial Business Intelligence and/or data analytics applications often offer rich visualization and interactive filtering. They combine a summary of information, with a drill-to-details option. The sentiment analysis application implemented for this project could be presented to its users not as a set of R Scripts producing static visualizations in PDF format, but rather as an interactive web application. The research conducted on technologies facilitating the creation of such interactive web application with R indicated that the Shiny framework[27] and the dygraphs R package[28] could be used as interfaces to the dygraphs JavaScript charting library[29] to improve interactivity.

The aggregate values of sentiment analysis could be presented in using a temporal dimension. In order to achieve that, text corpus would need to be enhanced by additional temporal attributes, and an append functionality. Temporal attributes could be created by extracting the date and time of tweet publications with a *$getCreated()* method of the twitteR package. Additionally, a script for downloading and processing the data would need to be slightly modified to append new tweets to an existing dataset. Finally, a method to schedule the load script would need to be introduced. On the Linux/Unix platform such functionality could be delivered by Cron, a system daemon for scheduling regular tasks in the background. For a

---

[22] https://github.com/pablobarbera/Rfacebook
[23] https://news.google.com/news
[24] http://www.interia.pl/
[25] http://www.onet.pl/
[26] http://www.wp.pl/
[27] http://shiny.rstudio.com/
[28] https://rstudio.github.io/dygraphs/
[29] http://dygraphs.com/

heterogenous text corpora containing data from twitter and other sources, a method to define a single approach to the time dimension across various data sources would need to be defined.

Assuming that the above idea is implemented, it would be practical to consider storing an increasing volume of data in a relational database. This functionality could be added into R by a package that embeds a self-contained, serverless, zero-configuration, transactional RDBMS database engine[30] called RSQLite[31].

Another idea is to enhance the dataset collection process to not only search for candidates, but also automatically extract hashtags from their twitter posts. Assuming that Presidential candidates are opinion leaders, their hashtags could be used to discover new forms of social activism on specific themes or content, which could contribute to a better text corpus used for Machine Learning, and building sentiment classifiers. There are a few such popular hashtags that were discovered, and could have been incorporated to the dataset. These were, on May 17th: #DebataPrezydencka, #CzasDecyzji, and later on May 21st, 2015: #debata, #debata2015 #debatatvn. All these popular topics commented on Twitter were related to the political debates of presidential candidates transmitted by Polish public and private TV channels.

To encourage other people to contribute to this project and/or to re-evaluate its findings, the author expects to publish (under the Apache 2.0 license) both the R code, and lexicons of positive/negative Polish words that were developed to produce results described in this paper.

---

[30] https://sqlite.org/
[31] http://cran.r-project.org/web/packages/RSQLite/index.html

# 6. Appendices

Other resources:
- 36 links to websites mentioned in page footnotes, all retrieved in April-June 2015, and valid as of June 17th, 2015.

## 6.2 List of Abbreviations

API(s) - Application Programming Interface(s)
BI - Business Intelligence
IDE - Integrated development environment
JSON - JavaScript Object Notation
LTS - Long Term Support
NLP - Natural language processing
NLU - Natural Language Understanding
OS - (Computer) Operating System
PPV - Positive Predictive Value
RDBMS - Relational Database Management System
RAM - Random Access Memory
TDM - Term-Document Matrix
TPR - True Positive Rate

## 6.3 Solutions to technical problems encountered

### 6.3.1 Configuration of development workstations

```
> sessionInfo()
R version 3.2.0 (2015-04-16)
Platform: i686-pc-linux-gnu (32-bit)
Running under: Ubuntu 14.04.2 LTS

locale:
 [1] LC_CTYPE=en_US.UTF-8       LC_NUMERIC=C               LC_TIME=pl_PL.UTF-8        LC_COLLATE=en_US.UTF-8
 [5] LC_MONETARY=pl_PL.UTF-8    LC_MESSAGES=en_US.UTF-8    LC_PAPER=pl_PL.UTF-8       LC_NAME=C
 [9] LC_ADDRESS=C               LC_TELEPHONE=C             LC_MEASUREMENT=pl_PL.UTF-8 LC_IDENTIFICATION=C

attached base packages:
[1] stats     graphics  grDevices utils     datasets  methods   base

other attached packages:
[1] twitteR_1.1.8 bnlearn_3.8.1 tm_0.6-1       NLP_0.1-7      stringi_0.4-1 ggplot2_1.0.1 dplyr_0.4.1

loaded via a namespace (and not attached):
 [1] Rcpp_0.11.6    magrittr_1.5   MASS_7.3-39    munsell_0.4.2  bit_1.1-12     colorspace_1.2-6 rjson_0.2.1
 [8] stringr_1.0.0  httr_0.6.1     plyr_1.8.2     tools_3.2.0    parallel_3.2.0 grid_3.2.0       gtable_0.1.
[15] DBI_0.3.1      bit64_0.9-4    assertthat_0.1 digest_0.6.8   reshape2_1.4.1 slam_0.1-32      scales_0.2.
[22] proto_0.3-10
> sessionInfo()
R version 3.1.3 (2015-03-09)
Platform: x86_64-w64-mingw32/x64 (64-bit)
Running under: Windows 7 x64 (build 7601) Service Pack 1

locale:
[1] LC_COLLATE=Polish_Poland.1250  LC_CTYPE=Polish_Poland.1250    LC_MONETARY=Polish_Poland.1250
[4] LC_NUMERIC=C                   LC_TIME=Polish_Poland.1250

attached base packages:
[1] stats     graphics  grDevices utils     datasets  methods   base

other attached packages:
[1] e1071_1.6-4       RTextTools_1.4.2  SparseM_1.6       tm_0.6-1          NLP_0.1-7         stringi_0.4-1    ggthemes_2.1.2
[8] ggplot2_1.0.1     dplyr_0.4.1

loaded via a namespace (and not attached):
 [1] assertthat_0.1   bitops_1.0-6     caTools_1.17.1   class_7.3-12     codetools_0.2-11
 [6] colorspace_1.2-6 DBI_0.3.1        digest_0.6.8     foreach_1.4.2    glmnet_2.0-2
[11] grid_3.1.3       gtable_0.1.2     ipred_0.9-4      iterators_1.0.7  lattice_0.20-30
[16] lava_1.4.0       lazyeval_0.1.10  magrittr_1.5     MASS_7.3-40      Matrix_1.1-5
[21] maxent_1.3.3.1   munsell_0.4.2    nnet_7.3-9       parallel_3.1.3   plyr_1.8.2
[26] prodlim_1.5.1    proto_0.3-10     randomForest_4.6-10 Rcpp_0.11.6    reshape2_1.4.1
[31] rpart_4.1-9      scales_0.2.4     slam_0.1-32      splines_3.1.3    stringr_0.6.2
[36] survival_2.38-1  tau_0.0-18       tools_3.1.3      tree_1.0-35
```

### 6.3.1 Unable to install TwitteR R package

```
installing to /home/griffi/R/i686-pc-linux-gnu-library/3.1/bit64/libs
** R
** data
** exec
** inst
** byte-compile and prepare package for lazy loading
in method for 'coerce' with signature '"character","integer64"': no definition for class "integer64"
in method for 'coerce' with signature '"integer64","character"': no definition for class "integer64"
** help
*** installing help indices
** building package indices
** testing if installed package can be loaded
* DONE (bit64)
ERROR: dependency 'RCurl' is not available for package 'httr'
* removing '/home/griffi/R/i686-pc-linux-gnu-library/3.1/httr'
Warning in install.packages :
  installation of package 'httr' had non-zero exit status
ERROR: dependency 'httr' is not available for package 'twitteR'
* removing '/home/griffi/R/i686-pc-linux-gnu-library/3.1/twitteR'
Warning in install.packages :
  installation of package 'twitteR' had non-zero exit status
```

**Problem description:** Two R packages identified as dependencies are not available. 'RCurl' is not available for package 'httr'.

**Remediation steps attempted:** Installing any of them independently did not help. Upgrading R packages from Ubuntu repository did not help. As described on the website of official CRAN repository, R packages for Ubuntu on i386 and amd64 are available for all stable Desktop releases of Ubuntu until their official end of life date. However, only the latest LTS release is fully supported[32]. As the author used the LTS release, he decided to update all R packages to their latest versions from the repository. Again, he did not achieve any progress in solving the problem.

**Solution:** The original thought was that the author encountered problems because he had fairly old version of R installed. He later discovered that the problem disappeared after he installed *libcurl* library and its dependent packages on the OS level. Libcurl is a is a free, open source client-side URL transfer library to get documents/files from servers, using any of the supported protocols[33]. It's available in official Debian and Ubuntu package repositories and can be installed by issuing the below commands:

```
griffi@Phantom:~$ sudo apt-get install libcurl
griffi@Phantom:~$ sudo apt-get install libcurl4-gnutls-dev
```

### 6.3.2 Warning messages in R console after starting RStudio

```
Error in tools:::httpdPort <= 0L :   comparison (4) is possible only for atomic and list types.
```

**Problem description:** The author always encountered the warning message (as per the above exhibit) just after starting RStudio. He used the RStudio version *0.98.1062 - Mozilla/5.0 (X11; Linux i686) AppleWebKit/534.34 (KHTML, like Gecko) RStudio Safari/534.34 Qt/4.8.0*.

**Remediation steps attempted:** He assumed that the above symptom is specific to RStudio IDE, because when started R in terminal window, the console did not produce the message.

---

[32] http://cran.r-project.org/bin/linux/ubuntu/

[33] Debian package description: https://packages.debian.org/sid/libdevel/libcurl4-gnutls-dev

**Solution:** This was a problem in the interaction between R and RStudio, which was solved by upgrading to the new version of RStudio. He downloaded the latest package for his platform and upgraded to version 0.98.1103. The warning message disappeared.

### 6.3.3 Unable to install XML R package

```
** package 'XML' successfully unpacked and MD5 sums checked
checking for gcc... gcc
checking for C compiler default output file name...
rm: cannot remove 'a.out.dSYM': Is a directory a.out
...
ERROR: configuration failed for package 'XML'
* removing '/home/griffi/R/i686-pc-linux-gnu-library/3.2/XML'
```

**Problem description:** The author considered to use XML R package for web scraping, to incorporate additional sources of data about Polish presidential elections into the text corpora. The problem encountered was related that the package did not install on his development workstation with install packages command: *install.packages("XML")*.

**Remediation steps attempted:** Upgrading R packages on OS level did not help.

**Solution:** It was discovered that the solution was to install one missing package called libxml2-dev[34] on his machine. The problem was fixed on OS level by issuing:

```
griffi@Phantom:~$                    sudo                    apt-get                    update
griffi@Phantom:~$ sudo apt-get install libxml2-dev
```

After the installation was complete, the install.packages("XML") command issued in R Console worked as expected, and the XML R package got compiled with no errors.

### 6.3.4 Twitter API Request Limits

Twitter clearly specified its API rules and provides guidelines to developers on how to avoid the data transfer rates being limited. These rules have to be considered when planning for any Twitter data collection. Data is accessible to registered Twitter users with two open and free API[35], which require authorization:
- REST API that provide programmatic access to read and write Twitter data.
- Streaming API to monitor or process Tweets in real-time.

These APIs are consumable in JSON or ATOM formats. This project uses REST API ver. 1.1 that is based on JSON format, - as it is a default method to connect to Twitter using twitteR package. API rate limits are defined per access token on a per-user and per-application basis. There are two versions of API, v1 and v1.1. Limits imposed depend on the version of API and on the request type. As a general rule the latest API allows for 15/30/180 requests (depending on the request type) per so called *rate limit window* of 15 minutes[36].

---

[34] As of May 14th, 2015: http://stackoverflow.com/questions/7765429/unable-to-install-r-package-in-ubuntu-11-04

[35] Twitter website for Developers - Documentation: https://dev.twitter.com/rest/public

[36] Twitter's website for Developers - API Rate Limits: https://dev.twitter.com/rest/public/rate-limits

### 6.3.5 Unable to process required number of tweets) using tm package

```
> corpus <- tm_map(raw.corpus, stripWhitespace, lazy=TRUE)
> dtm <- DocumentTermMatrix(corpus)
…
> freq <- colSums(as.matrix(dtm))
…
… Error: cannot allocate vector of size 1.1 Gb
```

**Problem description:** R is not able to allocate large enough contiguous block of address space. The size exceeded the address-space limit for a process or, the system was unable to provide the memory[37]. The author encountered memory related issues when building Document Term Matrix for his text corpora to explore its content and find frequent terms. This can be attributed to the limited number of optimization attempts and technical limitations of the development workstation (2 GB or RAM).

**Remediation steps attempted:**
1. Experimenting with parameter lazy=TRUE (so that the data is only processed if needed using so called lazy approach). This helped with executing *tm_map()* function in step one, but the overall processing was not successful.
2. Reducing the number of tweets in the dataset by half and reprocessing. The first successful attempt was processing 1468 tweets, so in theory one could design a sequential processing workflow. Attempt not tested due to limited time.

**Solution:** Assuming the time constraints the author decided to switch to a different development workstation, x64 machine with 8 GB RAM running Windows. He have not observed any memory related errors when working on that new machine.

```
System Manufacturer:         LENOVO
System Model:                20AMS2T101
System Type:                 x64-based PC
Processor(s):                1 Processor(s) Installed.
                             [01]: Intel64 Family 6 Model 69 Stepping 1 GenuineInt
el ~1875 Mhz
BIOS Version:                LENOVO GIET75WW (2.25 ), 2014-06-24
Windows Directory:           C:\WINDOWS
System Directory:            C:\WINDOWS\system32
Boot Device:                 \Device\HarddiskVolume1
System Locale:               en-us;English (United States)
Input Locale:                en-us;English (United States)
Time Zone:                   (UTC+01:00) Amsterdam, Berlin, Bern, Rome, Stockholm,
 Vienna
Total Physical Memory:       7 880 MB
Available Physical Memory:   1 415 MB
Virtual Memory: Max Size:    15 758 MB
Virtual Memory: Available:   9 250 MB
Virtual Memory: In Use:      6 508 MB
Page File Location(s):       C:\pagefile.sys
```

## 6.4 2015 Polish Presidential Election on Twitter

### 6.4.1 Candidates in the 1st Round of Presidential Election

As of April 12th, 2015

---

[37] http://stackoverflow.com/questions/5171593/r-memory-management-cannot-allocate-vector-of-size-n-mb

| Candidate's Name & Surname | Twitter Account | Tweets & Followers | Bot or Not Score *) |
|---|---|---|---|
| Duda Andrzej Sebastian | @AndrzejDuda2015 <br> @AndrzejDuda | 971 \| 4979 <br> 5918 \| 26977 | 41% \| 13% <br> 22% \| 21% |
| Jarubas Adam Sebastian | @JarubasAdam | 914 \| 3557 | 33% \| 11% |
| Bronisław Maria Komorowski | @Komorowski | 540 \| 26771 | 26% \| 37% |
| Korwin-Mikke Janusz Ryszard | @JkmMikke <br> @korwinmikke | 54 \| 6405 <br> 3366 \| 33911 | 42% \| 40% <br> 31% \| 34% |
| Kowalski Marian Janusz | @M_Kowalski1 | 193 \| 2500 | 40% \| 16% |
| Kukiz Paweł Piotr | @PrezydentKukiz <br> @pkukiz <br> @KukizPawelKukiz | 994 \| 6878 <br> 25 \| 1517 <br> 10 \| 2089 | 37% \| 29% <br> 43% \| 20% <br> 54% \| 79% |
| Ogórek Magdalena Agnieszka | @ogorekmagda | 557 \| 11834 | 29% \| 6% |
| Palikot Janusz Marian | @Palikot_Janusz | 6453 \| 347873 | 23% \| 19% |
| Paweł Jan Tanajno | @Pawel_Tanajno | 16 \| 42 | 29% \| 48% |
| Wilk Jacek | @JacekWilkPL | 319 \| 2996 | 31% \| 19% |
| Grzegorz Michał Braun | N/A | N/A | N/A \| N/A |

*) Bot or Not score acquired from 'A Truthy project': http://truthy.indiana.edu/botornot/

## 6.4.2 Candidates in the 2st Round of Presidential Election

As of May 20th, 2015

| Candidate's Name & Surname | Twitter Account | Tweets & Followers | Bot or Not Score *) |
|---|---|---|---|
| Bronisław Maria Komorowski | @Komorowski | 1277 (+737) <br> 34634 (+7863) | 26% \| 37% |
| Andrzej Sebastian Duda | @AndrzejDuda2015 | 1851 (+880) <br> 9335 (+4356) | 41% \| 13% |
|  | @AndrzejDuda | 6470 (+552) <br> 38408 (+11431) | 22% \| 21% |

*) Bot or Not score acquired from 'A Truthy project': http://truthy.indiana.edu/botornot/

Tweet Sentiment Visualization using the tool from [10], random tweet sample for @komorowski and @AndrzejDuda

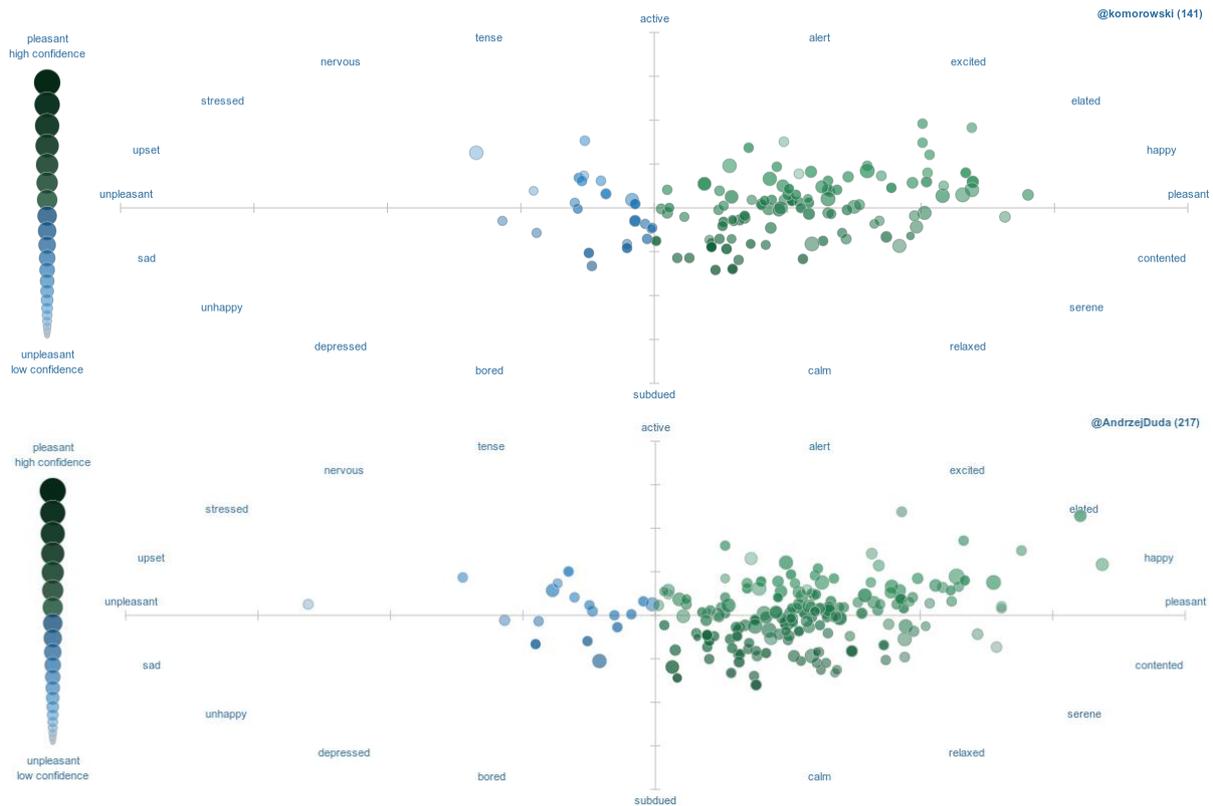